\documentclass[10pt, conference, compsocconf]{IEEEtran}
\usepackage{times}
\usepackage{epsfig}
\usepackage{graphicx}
\usepackage{amsmath}
\usepackage{amssymb}
\usepackage[tight,footnotesize]{subfigure}
\usepackage{cite}
\usepackage{url}
\usepackage{subfig}
\usepackage{caption}
\usepackage{array}
\ifCLASSINFOpdf
\else
\fi
\begin{document}
%
\title{A Bag of Words Approach for Semantic Segmentation of Monitored Scenes}
\author{\IEEEauthorblockN{Wassim Bouachir, Atousa Torabi, and Guillaume-Alexandre Bilodeau}
\IEEEauthorblockA{LITIV Lab.\\
\'Ecole Polytechnique de Montr\'eal\\
Montr\'eal (Qu\'ebec), Canada\\
\{wassim.bouachir, atousa.torabi, gabilodeau\}@polymtl.ca}
\and
\IEEEauthorblockN{Pascal Blais}
\IEEEauthorblockA{Department of Research and Development\\
iWatchLife Inc.\\
Ottawa (Ontario), Canada\\
pascal.blais@iwatchlife.com}
}
\maketitle
\begin{abstract}
This paper proposes a semantic segmentation method for outdoor scenes captured by a surveillance camera. Our algorithm classifies each perceptually homogenous region as one of the predefined classes learned from a collection of manually labelled images. The proposed approach combines two different types of information. 
First, color segmentation is performed to divide the scene into perceptually similar regions. Then, the second step is based on SIFT keypoints and uses the bag of words representation of the regions for the classification. The prediction is done using a Na\"ive Bayesian Network as a generative classifier. Compared to existing techniques, our method provides more compact representations of scene contents and the segmentation result is more consistent with human perception due to the combination of the color information with the image keypoints. The experiments conducted on a publicly available data set demonstrate the validity of the proposed method.
\end{abstract}
%
\IEEEpeerreviewmaketitle
\section{Introduction}
Semantic image segmentation is the task of partitioning an image into a number of meaningful regions. This task is different from low-level segmentation, the ultimate goal being to assign the correct class to image regions rather than simply finding objects boundaries. A variety of applications, such as object detection, image annotation, and content-based indexing use semantic segmentation algorithms. In our work, we propose a semantic scene segmentation algorithm for an intelligent video surveillance system in outdoor environments. Segmenting and identifying the different regions in a monitored scene is a very useful preprocessing step for detecting and recognizing objects and events of interest. For example, a car should ride on a road, not in the sky. Knowing tree regions also allows identifying regions where background subtraction may fail, like in the case of swaying trees. 

In the literature, a large number of image features and induction methods have been proposed for segmenting and labelling images. In many works \cite{berg2007, robin2007,gould2008, kohli2009, vez2011}, the authors used super-pixels characterized by one or more features such as color, texture, dominant orientations, and shape. For region recognition tasks, several induction models use global image features, local visual features, or a combination of both. In \cite{jiebo2001}, the authors use a Bayesian Network that combines local and global features to classify indoor and outdoor images. The method proposed in \cite{sang2001} uses a hierarchical clustering procedure to segment the image, and then estimates the classes of the segmented regions by Fuzzy classification. In addition to image content, another approach integrates the temporal context to exploit the dependence between successive images captured within a short period of time \cite{boutel2004}.
Generally, the most prevalent model used for image segmentation is the Markov Random Field \cite{9,10,11,Liu2010,saxena9, saxena5}. 

In our work, we deal with the same problem of outdoor semantic image segmentation as in \cite{11}, but with a different approach. In \cite{11}, the segmentation of street view images is obtained by a Markov Random Field model. Color and shape features are extracted at a super-pixel level, and recognition is then done by multiple adaBoost classifiers \cite{12}. Most of existing techniques suffer from high computational complexity for training a predictor, segmenting the regions, and predicting the labels. Moreover, the major limitation of Markov Random Field segmentation is that its complexity increases considerably with large images.
In this paper, we present a novel semantic labelling method to partition an input image into a number of non-overlapping and meaningful regions. The proposed algorithm is designed to be integrated to an intelligent video surveillance system. We aim to use our method in visual surveillance, and label images regularly to validate detections and events. The low-level partitioning is produced by a fast graph-based color segmentation method. The labelling of each obtained region is induced from a set of manually segmented images. For this purpose, a Na\"ive Bayesian classifier is constructed using a training set. 

The main contributions of this paper are: 1) a compact representation of scene contents that combines keypoint features with color statistics, and 2) a simple and efficient classification model to recognize the image regions in a supervised approach. 
The paper is organized as follows: the second section describes the bag of words framework. Sections III and IV respectively present the proposed classification model and the color segmentation method that we use in our work. Section V provides detailed experimental results, and section VI concludes the paper.
\section {The bag of words indexing}
Since its introduction by Sivic and Zisserman in \cite{13}, the bag of words model has been widely used in computer vision tasks thanks to its robustness and low computational complexity \cite{14,15,16}. This model describes each image segment using a set of visual patterns called visual vocabulary. The vocabulary is obtained by clustering local features extracted from manually segmented images, where each resulting cluster is a visual word. An image segment is finally represented by a histogram. Each bin of this histogram corresponds to a visual word, and the associated weight represents its importance in the segment. The construction of the histogram requires three steps: 1) extracting local features, 2) building a visual vocabulary, and 3) creating signatures.
\subsection{Extracting local features}
To extract the local features from image segments, we detect keypoints. Keypoints are the centers of salient patches generally located around the corners and edges. In our work, we detect and describe keypoints using the Scale Invariant Features Transform (SIFT) \cite{17}. Our method is not specific to SIFT. Even faster keypoint detector/descriptor combination may be used, although SIFT remains one of the most reliable method under various image transformations \cite{heinly12}. In this step, SIFT keypoints are extracted from manually segmented image regions, and each keypoint is described by a vector of 128 elements summarizing local gradient information. The extracted features will be used to build the visual vocabulary.
\subsection{Building the visual vocabulary}
Building the visual vocabulary means quantifying extracted local descriptors. The vocabulary is generated by clustering SIFT features using the standard k-means algorithm. The size of the vocabulary is the number of clusters, and the centers of the clusters are the visual words. Each image segment in the database will be represented by visual words from this vocabulary.
\subsection{Creating signatures}
Once the visual vocabulary is built, we index each segment by constructing its bag of words signature. This requires finding the weight of the visual words from the vocabulary. Each segment is described by a histogram, where the k bins are the visual words and the corresponding values are the weights of the words in the image region. To compute the weight of a visual word of a given region, we apply the fuzzy weighting scheme proposed in \cite{18} and defined by the equation:
\begin{equation}
U_{ij} = \frac{1}
{
\sum_{n=1}^k (\frac{\mid\mid p_{j}-v_{i}\mid\mid}{\mid\mid p_{j}-v_{n}\mid\mid} )^{\frac{2}{m-1}}
}
\end{equation}
where $U_{ij}$ is the contribution of the keypoint described by the feature vector $p_{j}$ in the weight of the visual word $v_{i}$, and $m$ is the degree of fuzziness. This weighting scheme maintains the simplicity and efficiency of the bag of words approach, while producing a fuzzy signature that reflects the real weights of the visual words. 

Since SIFT  features are based on gradient orientation histograms computed on grayscale images, we extend the signature by the RGB color information. The mean and the variance of the three channels are computed for each region and added to form an extended bag of words signature.
\section{The Bayesian classification model}
The Na\"ive Bayesian Network has been widely used for bag of words text categorization because of its simplicity, learning speed and competitiveness with the state-of-the art classifiers\cite{csurka2004}. In our work, we propose to use the Na\"ive Bayesian Network as a generative classifier to learn a set of classes and recognize the regions of an input image. While attribute independence is a na\"ive assumption, the accuracy of the Na\"ive Bayes classification is proven to be typically high\cite{domingos1997}. The main idea of our model is to learn from a training set the conditional probability of each attribute given a class. The classification decision is taken by applying Bayes’ rule:
\begin{equation}
P(C_{i}|X_{n}) = \frac{P(C_{i})P(X_{n}|C_{i})}
{ P(X_{n}) }
\end{equation}
where $P(C_{i} |X_{n})$ is the probability of the category $C_{i}$ given $X_{n}$ (the signature of a segment $S_{n}$). $P(C_{i})$ and $P(X_{n} )$ are respectively the prior probability of the class $C_{i}$, and the prior probability of obtaining the signature $X_{n}$ for a segment. The probability $P(X_{n})$ is the same for all the classes, and therefore, it can be ignored without affecting the relative values of class probabilities. Finally, we consider the largest a posteriori score as the class prediction.

 The segments signatures contain real valued attributes that represent the weights of visual words and the color statistics. To model the conditional probabilities distributions, we assume that for a given class $C_{i}$, the feature $w_{j}$ is a normally distributed random variable with mean $\mu_{ij}$ and variance $\sigma_{ij}^2$. After decomposing $P(X_{n} |C_{i})$ into the product of the conditional probabilities learned for each attribute value we obtain:
\begin{equation}
P(C_{i}|X_{n}) = P(C_{i}) \prod_{j=1}^k P(w_{j}=v|C_{i}).
\end{equation}
The \textit {a posteriori} score of classes is then computed using equation (3) with:
\begin{equation}
P(w_{j}=v|C_{i}) = \frac{1} {\sqrt{2\pi} \sigma_{ij}} e^
{
-\frac{(v-\mu_{ij})^2}{2\sigma_{ij}^2}
}
\end{equation}
where $v\in [0;\infty[$.
The proposed classifier uses a training set of signatures corresponding to manually segmented images. To predict the classes for the different regions in a new scene, we need first to segment the image automatically.
\section{Image segmentation}
To segment an unknown input image into a set of perceptually homogeneous regions, we used the color segmentation algorithm proposed by Felzenszwalb and Huttenlocher \cite{19}. This graph-based method has been widely used for automatic segmentation and semantic ROI classification. It incrementally merges regions of similar appearance with small minimum spanning tree weight. This method has nearly linear computational complexity which allows it to be fast in practice, running in a fraction of second. The other important characteristic is its ability to preserve details in low-variability image regions, while ignoring details in high-variability regions. We note here that our framework is not tied to this segmentation algorithm. Any method that would provide a reasonable segmentation of the scenes would suit our need. Once the input image is decomposed into a set of homogeneous regions, each region is indexed by constructing its bag of words signature. The classification task is finally performed by applying the Na\"ive Bayesian classifier on the obtained signatures as explained in the previous section.
\section{Experiments}
The evaluation of the proposed method was conducted on the public data set\footnote{The data set is available at: http://dags.stanford.edu/projects/scenedataset} used by the authors in \cite{Liu2010, saxena9, saxena5}. This data set contains 534 outdoor images of size 240x320 pixels. The collection is divided into a training set of 400 images and an evaluation set of 134 images, where each scene is labelled into the semantic classes: sky, tree, road, grass and building. Figure 1 shows sample images from the data set. 
\begin{figure*}
\centering
\subfigure{
\includegraphics[width=4cm]{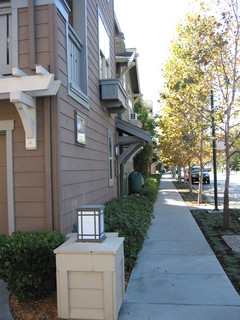}}
\subfigure{
\includegraphics[width=4cm]{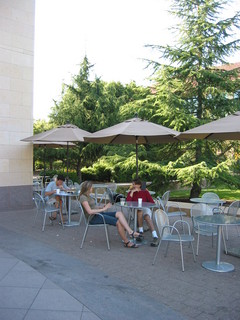}}
\subfigure{
\includegraphics[width=4cm]{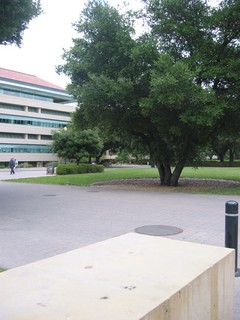}}
\subfigure{
\includegraphics[width=4cm]{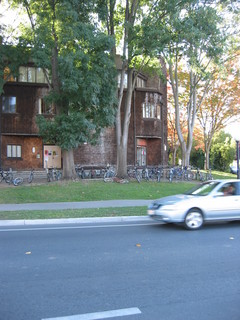}}
\caption[example] 
{ \label{fig:example} 
Sample images from the data set. }
\end{figure*}
We extracted SIFT keypoints from the training set and we used the k-means clustering algorithm to cluster the extracted local features into a visual vocabulary. For our experiments, we set the size of the vocabulary to 60 visual words. This vocabulary size gives the best results experimentally, as it is a good compromise between precision and generality. The manually segmented regions are then indexed by computing their bag of words signatures. Once the training set images are indexed, the obtained signatures are used to train the Na\"ive Bayesian classifier.

Given an unknown input image from the evaluation set, we first apply the graph-based segmentation algorithm. We then extract SIFT keypoints and color information to compute, for each region, the corresponding signature considering the visual vocabulary. For a given region of an image, the class recognition is finally done by applying the Bayes’ rule defined in equation (3). The metric that we are using for evaluation is the classification rate, which is defined as the ratio between the number of correctly classified regions for a given class, and the number of regions belonging to that class. 
Examples of semantic segmentation results obtained for three evaluation images are shown in figure 2, illustrating the performance of the proposed approach. Note that our recognition method is independent of the color segmentation algorithm. As a consequence, even if the segmentation algorithm may divide a semantically homogeneous region into two or more segments based on color features, often the classification algorithm assigns the same correct label to all the segments as we can see in figure 2-d.  In fact, over-segmentation is preferable to under-segmentation as only one label is assigned per region, and thus the segmentation method should be configured for slight over-segmentation.
\begin{figure*}
\centering
\subfigure[]{
\includegraphics[width=5cm]{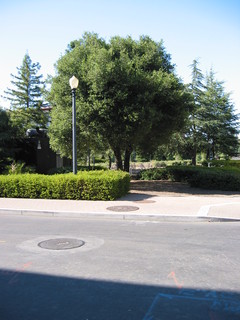}}
\subfigure[]{
\includegraphics[width=5cm]{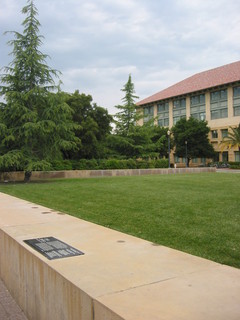}}
\subfigure[]{
\includegraphics[width=5cm]{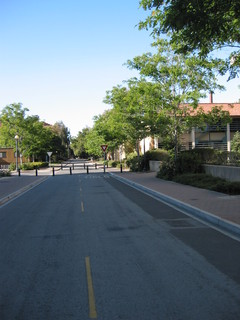}}
\subfigure[]{
\includegraphics[width=5cm]{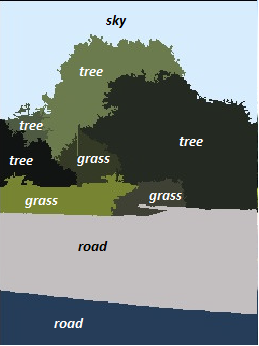}}
\subfigure[]{
\includegraphics[width=5cm]{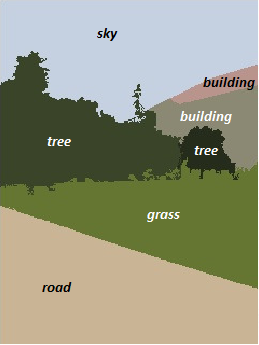}}
\subfigure[]{
\includegraphics[width=5cm]{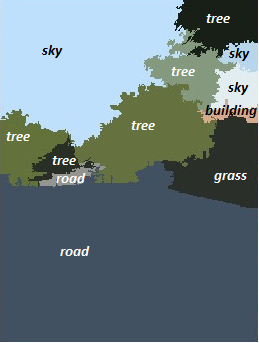}} 
\caption[example] 
{ \label{fig:example} 
Semantic segmentation on test images using the bag of words representation extended by color statistics, and the Na\"ive Bayesian classifier.}
\end{figure*}

The confusion matrix obtained by using only the bag of words signatures (without color statistics) for the 134 evaluation images is given in table I. In this table, the average classification rate is 72\% and the diagonal elements show high classification rates for most of the classes. 
\begin{table}
\renewcommand {\arraystretch }{1.5}
\centering
\begin{tabular}{| l | c c c c c |}
\hline
$\downarrow$ \textit{True classes} & \textit{Sky} & \textit{Tree} & \textit{Road} & \textit{Grass} & \textit{Building} \\
\hline\hline
\textit{Sky} & \textbf {86} & 1 & 2 &2 &9 \\
\textit{Tree} & 0 & \textbf {87} & 0 &8 &5 \\
\textit{Road} & 2 & 13 & \textbf {72} & 8 & 5 \\
\textit{Grass} & 1 &20 &26& \textbf {40}& 13\\
\textit{Building} & 6& 5& 14& 9& \textbf {66}\\
\hline
\end{tabular}
\caption { Confusion matrix (in percentages) for the classification results using the bag of words representation (without the color statistics) and the Na\"ive Bayesian classifier.}
\end{table}
Furthermore, we extend the bag of words signature by the mean color and the variance of each channel (R, G, and B) in the indexed region. Table II shows that adding color statistics to the bag of words signature increases the average correct classification rate from 72\% to 80\%. It also shows a high classification rate of 88\% for classes sky and tree. The lowest classification rate is 74\%, and it was obtained for classes road and building. This percentage can be explained by the fact that objects of confused categories can be similar in appearance. For example, 11\% of building regions were confused with the category road because segments of these two classes can be similar in texture (see figure 2-e and 2-f).
\begin{table}
\renewcommand {\arraystretch }{1.5}
\centering
\begin{tabular}{| l | c c c c c |}
\hline
$\downarrow$ \textit{True classes} & \textit{Sky} & \textit{Tree} & \textit{Road} & \textit{Grass} & \textit{Building} \\
\hline\hline
\textit{Sky} & \textbf {88} & 1 & 1 &2 &8 \\
\textit{Tree} & 0 & \textbf {88} & 0 &6 &6 \\
\textit{Road} & 4 & 10 & \textbf {74} & 7 & 5 \\
\textit{Grass} & 0 &21 &0& \textbf {76}& 3\\
\textit{Building} & 4& 3& 11& 8& \textbf {74}\\
\hline
\end{tabular}
\caption {Confusion matrix (in percentages) for the classification results using the bag of words representation (extended by the color statistics) and the Na\"ive Bayesian classifier.}
\end{table}
\section{Conclusion and future work}
We proposed a novel semantic segmentation method to enable efficient labelling of images captured by a surveillance camera. A state-of-the-art segmentation algorithm is firstly used to find the boundaries of homogeneous regions based on color. The proposed recognition method relies on a bag of words representation extended by color statistics for visual indexing. The classification decisions for image regions are taken by a Na\"ive Bayesian classifier trained on a data set of manually segmented images.

Our future work will be focused on how to incrementally improve the recognition results by adding a feedback procedure for the video surveillance system user. In more concrete terms, when a semantic segmentation is performed for a new scene, the user would be able to select one or several correctly classified regions. The return of this information would allow updating the classifier in order to improve future performances.
\section*{Acknowledgment}
This work is supported by the Natural Sciences and Engineering Research Council of Canada (NSERC).
\bibliographystyle{IEEEtran}
\bibliography{references}
%
%
%
\end{document}